%% file: main.tex
\documentclass[11pt,a4paper]{article}
\usepackage[hyperref]{emnlp2018}
\usepackage{times}
\usepackage{latexsym}
\usepackage{graphicx}

\usepackage{url}

\aclfinalcopy 

\usepackage{booktabs}
\usepackage{array}
\usepackage{adjustbox}
\usepackage[T1]{fontenc}
\usepackage{float}
\usepackage{subcaption}
\usepackage{algpseudocode}
\usepackage{algorithm}
\usepackage{caption}
\usepackage{amsmath,amssymb}

\usepackage[colorinlistoftodos]{todonotes}

\usepackage{multirow}

\input{defs}

\title{Joint Multitask Learning for Community Question Answering\\ Using Task-Specific Embeddings}

\author{Shafiq Joty$^\dagger$, Llu\'{i}s M\`{a}rquez\thanks{\vspace{5pt} Work conducted while this author was at QCRI, HBKU.} \  and Preslav Nakov$^\ddagger$ \\
  $^\dagger$Nanyang Technological University, Singapore \\
  $^\star$Amazon, Barcelona, Spain\\
  $^\ddagger$Qatar Computing Research Institute, HBKU, Qatar\\
  {\tt srjoty@ntu.edu.sg,   lluismv@amazon.com,   pnakov@qf.org.qa} 
}

\date{}

\begin{document}
\maketitle
\begin{abstract}

We address jointly two important tasks for Question Answering in community forums: given a new question, \Ni \emph{find related existing questions}, and \Nii \emph{find relevant answers to this new question}.  We further use an auxiliary task to complement the previous two, i.e.,~\Niii \emph{find good answers with respect to the thread question in a question-comment thread}. We use deep neural networks (DNNs) to learn meaningful task-specific embeddings, which we then incorporate into a conditional random field (CRF) model for the multitask setting, performing joint learning over a complex graph structure. While DNNs alone achieve competitive results when trained to produce the embeddings, the CRF, which makes use of the embeddings and the dependencies between the tasks, improves the results significantly and consistently across a variety of evaluation metrics, thus showing the complementarity of DNNs and structured learning.

\end{abstract}

\section{Introduction and Motivation}
\input{introduction}

\label{sec:intro}

\section{Related Work}
\input{related}
\label{sec:related}

\section{Learning Approach}
\label{sec:approach}	
\input{models-intro}

\subsection{Neural Models for cQA Subtasks}
\label{subsec:dnn-model}
\input{dnn-model}

\subsection{Joint Learning with Global Inference}
\label{subsec:crf-model}
\input{crf-model }

\section{Features for the DNN Models}
\label{sec:features}
\input{features}

\section{Data and Settings}
\label{sec:data-setting}
\input{data-setting}

\section{Results and Discussion}
\label{sec:results}

Below, we first present the evaluation results using DNN models (Section~\ref{subsec:dnn-eval}).
Then, we discuss the performance of the joint models (Section~\ref{subsec:joint-eval}).

\subsection{Results for the DNN Models}
\label{subsec:dnn-eval}
\input{dnn-eval}

\subsection{Results for the Joint Model}
\label{subsec:joint-eval}
\input{joint-eval}

\section{Conclusion}
\label{sec:conclusion}

We have presented a framework for multitask learning of two community Question Answering problems: question-question relatedness and answer selection. We further used a third, auxiliary one, i.e., finding the good comments in a question-comment thread. We proposed a two-step framework based on deep neural networks and structured conditional models, with a feed-forward neural network to learn task-specific embeddings, which are then used in a pairwise CRF as part of a multitask model for all three subtasks. 

The DNN model has its strength in generating compact embedded representations for the subtasks by modeling interactions between different input elements.

\noindent On the other hand, the CRF is able to perform global inference over arbitrary graph structures accounting for the dependencies between subtasks to provide globally good solutions. The experimental results have proven the suitability of combining the two approaches. The DNNs alone already yielded competitive results, but the CRF was able to exploit the task-specific embeddings and the dependencies between subtasks to improve the results consistently across a variety of evaluation metrics, yielding state-of-the-art results.

In future work, we plan to model text complexity \cite{SemEval2016:task3:SUper}, veracity \cite{AAAI2018:factchecking}, speech act \cite{joty-hoque:2016}, user profile \cite{mihaylov-georgiev-nakov:2015:CoNLL}, trollness
\cite{InternetResearchJournal:2018}, and goodness polarity \cite{SemEval2016:task3:PMI-cool,PMI:SIGIR:2017}. 
From a modeling perspective, we want to strongly couple CRF and DNN, so that the global errors are backpropagated from the CRF down to the DNN layers.  It would be also interesting to extend the framework to a cross-domain \cite{EMNLP2018:adversarial} or a
cross-language setting \cite{DaSanMartino:2017:CQR,joty-EtAl:2017:CoNLL}. Trying an ensemble of neural networks with different initial seeds  is another possible research direction.

\section*{Acknowledgments}

The first author would like to thank the funding support from MOE Tier-1. 

\bibliography{tacl2016}
\bibliographystyle{acl_natbib_nourl}

\end{document}

%% file: defs.tex
\DeclareMathOperator*{\Ber}{Ber}
\DeclareMathOperator*{\sig}{sig}
\DeclareMathOperator \expect{\mathbb{E}}

\newcommand{\Ni}{({\em i})~}
\newcommand{\Nii}{({\em ii})~}
\newcommand{\Niii}{({\em iii})~}

\newcommand{\Na}{({\em A})~}
\newcommand{\Nb}{({\em B})~}
\newcommand{\Nc}{({\em C})~}

%% file: introduction.tex

Question answering web forums such as StackOverflow, Quora, and Yahoo! Answers usually organize their content in topically-defined forums containing multiple \emph{question--comment threads}, where a question posed by a user is often followed by a possibly very long list of comments by other users, supposedly intended to answer the question. Many forums are not moderated, which often results in noisy and redundant content.

Within community Question Answering (cQA) forums, two subtasks are of special relevance when a user poses a new question to the website \cite{cQA:Survey:2018,lai-bui-li:2018:C18-1}: \Ni finding similar questions (\emph{question-question relatedness}), and \Nii finding relevant answers to the new question, if they already exist (\emph{answer selection}).

\noindent Both subtasks have been the focus of recent research as they result in end-user applications. The former is interesting for a user who wants to explore the space of similar questions in the forum and to decide whether to post a new question. It can also be relevant for the forum owners as it can help detect redundancy, eliminate question duplicates, and improve the overall forum structure. Subtask \Nii on the other hand is useful for a user who just wants a quick answer to a specific question, without the need of digging through the long answer threads and winnowing good from bad comments or without having to post a question and then wait for an answer.%

Obviously, the two subtasks are interrelated as the information needed to answer a new question is usually found in the threads of highly related questions. Here, we focus on jointly solving the two subtasks with the help of yet another related subtask, i.e., determining whether a comment within a question-comment thread is a good answer to the question heading that thread.  

An example is shown in Figure~\ref{fig:example}.  A new question $q$ is posed for which several potentially related questions are identified in the forum (e.g., by using an information retrieval system); $q_i$ in the example is one of these existing questions. Each retrieved question comes with an associated thread of comments; $c_m^i$ represents one comment from the thread of question $q_i$. 
Here, $c_m^i$ is a \emph{good} answer for $q_i$, $q_i$ is indeed a question \emph{related} to $q$, and consequently $c_m^i$ is a \emph{relevant} answer for the new question $q$. This is the setting of SemEval-2016 Task 3, and we use its benchmark datasets.

\begin{figure}[t]
\small
\begin{itemize}

\item[$q$:] ``{\bf How can I extend a family visit visa?}''
%
\item[$q_i$:] ``Dear All; I wonder if anyone knows the procedure how I can extend the family visit visa for my wife beyond 6 months. I already extended it for 5 months and is 6 months running. I would like to get it extended for couple of months more.Any suggestion is highly appreciable.Thanks''
%
\item[$c_m^i$:] ``You can get just another month's extension before she completes 6 months by presenting to immigration office a confirmed booking of her return ticket which must not exceed 7 months.''
\end{itemize}
\vspace{-0.5em}
\caption{\label{fig:example}Example of the three pieces of information in the cQA problems addressed in this paper.}
\vspace{-0.5em}
\end{figure}

Our approach has two steps. First, a deep neural network (DNN) in the form of a feed-forward neural network is trained to solve each of the three subtasks separately, and the subtask-specific hidden layer activations are taken as embedded feature representations to be used in the second step.

\noindent Then, a conditional random field (CRF) model uses these embeddings and performs joint learning with global inference to exploit the dependencies between the subtasks. 

A key strength of DNNs is their ability to learn nonlinear interactions between underlying features through specifically-designed hidden layers, and also to learn the features (e.g., vectors for words and documents) automatically. This capability has led to gains in many unstructured output problems. DNNs are also powerful for structured output problems. Previous work has mostly relied on recurrent or recursive architectures to propagate information through hidden layers, but has been disregarding the modeling strength of structured conditional models, which use global inference to model consistency in the output structure (i.e., the class labels of all nodes in a graph). In this work, we explore the idea that combining simple DNNs with structured conditional models can be an effective and efficient approach for cQA subtasks that offers the best of both worlds.

Our experimental results show that: \Ni DNNs already perform very well on the question-question similarity and answer selection subtasks; \Nii strong dependencies exist between the subtasks under study, especially answer-goodness and question-question-relatedness influence answer-selection significantly; \Niii the CRFs exploit the dependencies between subtasks, providing sizeably better results that are on par or above the state of the art. In summary, we demonstrate the effectiveness of this marriage of DNNs and structured conditional models for cQA subtasks, where a feed-forward DNN is first used to build vectors for each individual subtask,
which are then ``reconciled'' in a multitask CRF.

%% file: related.tex

Various neural models have been applied to cQA tasks such as  \emph{question-question similarity} \cite{dossantos-EtAl:2015,lei-EtAl:2016:N16-1,Wang:2018:CAC:3183892.3151957} and \emph{answer selection} ~\cite{wang-nyberg:2015:ACL-IJCNLP,qiu2015convolutional,tan2015lstm,chen-bunescu:2017:I17-2,wu-sun-wang:2018:Long}. Most of this work used advanced neural network architectures based on convolutional neural networks (CNN), long short-term memory (LSTM) units, attention mechanism, etc. 
For instance, \newcite{dossantos-EtAl:2015} combined CNN and bag of words for comparing questions. \newcite{tan2015lstm} adopted an attention mechanism over bidirectional LSTMs to generate better answer representations, and \newcite{lei-EtAl:2016:N16-1} combined recurrent and CNN models for question representation. 
In contrast, here we use a simple DNN model, i.e., a feed-forward neural network, which we only use to generate task-specific embeddings, and we defer the joint learning with global inference to the structured model.

From the perspective of modeling cQA subtasks as structured learning problems, there is a lot of research trying to exploit the correlations between the comments in a question--comment thread. This has been done from a feature engineering perspective, by modeling a comment in the context of the entire thread \cite{barroncedeno-EtAl:2015:ACL-IJCNLP}, but more interestingly by considering a thread as a structured object, where comments are to be classified as \emph{good} or \emph{bad} answers collectively. 
For example, \newcite{zhou-EtAl:2015:ACL-IJCNLP} treated the answer selection task as a sequence labeling problem and used recurrent convolutional neural networks and LSTMs.
\newcite{joty:2015:EMNLP} modeled the relations between pairs of comments at any distance in the thread, and combined the predictions of local classifiers using graph-cut and Integer Linear Programming.
In a follow up work, \newcite{joty-marquez-nakov:2016:N16-1} also modeled the relations between all pairs of comments in a thread, but using a fully-connected pairwise CRF model, which is a joint model that integrates inference within the learning process using global normalization. Unlike these models, we use DNNs to induce task-specific embeddings, and, more importantly, we perform multitask learning of three different cQA subtasks, thus enriching the relational structure of the graphical model.

\noindent We solve the three cQA subtasks jointly, in a multitask learning framework. We do this using the datasets from the SemEval-2016 Task~3 on \emph{Community Question Answering}~\cite{nakov-EtAl:2016:SemEval}, which are annotated for the three subtasks, and we compare against the systems that participated in that competition. In fact, most of these systems did not try to exploit the interaction between the subtasks or did so only as a pipeline. For example, the top two systems, \textsc{SUper team} \cite{SemEval2016:task3:SUper} and \textsc{Kelp}~\cite{filice-EtAl:2016:SemEval}, stacked the predicted labels from two subtasks in order to solve the main answer selection subtask using SVMs. 
In contrast, our approach is neural, it is based on joint learning and task-specific embeddings, and it is also lighter in terms of features.

In work following the competition, \newcite{nakov-marquez-guzman:2016:EMNLP2016} used a triangulation approach to answer ranking in cQA, modeling the three types of similarities occurring in the triangle formed by the original question, the related question, and an answer to the related comment. However, theirs is a pairwise ranking model, while we have a joint model. Moreover, they focus on one task only, while we use multitask learning.
\newcite{bonadiman-uva-moschitti:2017:EACLshort} proposed a multitask neural architecture where the three tasks are trained together with the same representation. However, they do not model comment-comment interactions in the same question-comment thread nor do they train task-specific embeddings, as we do.

The general idea of combining DNNs and structured models has been explored recently for other NLP tasks. \newcite{collobert2011natural} used Viterbi inference to train their DNN models to capture dependencies between word-level tags for a number of sequence labeling tasks: ~part-of-speech tagging, chunking, named entity recognition, and semantic role labeling. \newcite{HuangXY15} proposed an LSTM-CRF framework for such tasks. \newcite{ma-hovy:2016:P16-1} included a CNN in the framework to compute word representations from character-level embeddings. While these studies consider tasks related to constituents in a sentence, e.g.,~words and phrases, we focus on methods to represent comments and to model dependencies between comment-level tags. We also experiment with arbitrary graph structures in our CRF model to model dependencies at different levels.   

%% file: models-intro.tex

Let $q$ be a newly-posed question, and $c_m^i$ denote the $m$-th comment $(m\in \{1, 2, \ldots, M \})$ in the answer thread for the $i$-th potentially related question $q_i$ $(i\in \{1, 2, \ldots, I\})$ retrieved from the forum. We can define three cQA subtasks: \Na classify each comment $c_m^i$ in the thread for question $q_i$ as \emph{Good} vs. \emph{Bad} with respect to $q_i$; \Nb determine, for each retrieved question $q_i$, whether it is \emph{Related} to the new question $q$ in the sense that a good answer to $q_i$ might also be a good answer to $q$; and finally, \Nc classify each comment $c_m^i$ in each answer thread as either \emph{Relevant} or \emph{Irrelevant} with respect to the new question $q$.

Let $y_{i,m}^a$ $\in$ $\{Good, Bad\}$, $y_{i}^b$ $\in$ $\{Related,$ $Not$-$related\}$, and $y_{i,m}^c \in \{Relevant, Irrelevant\}$ denote the corresponding output labels for subtasks A, B, and C, respectively. As argued before, subtask C depends on the other two subtasks. Intuitively, if $c_m^i$ is a good comment with respect to the existing question $q_i$, and $q_i$ is related to the new question $q$ (subtask A), then $c_m^i$ is likely to be a relevant answer to $q$.
Similarly, subtask B can benefit from subtask C: if comment $c_m^i$ in the answer thread of $q_i$ is relevant with respect to $q$, then $q_i$ is likely to be related to $q$. 

We propose to exploit these inherent correlations between the cQA subtasks as follows: \Ni by modeling their interactions in the input representations, i.e., in the feature space of $(q, q_i, c_m^i)$, and more importantly, \Nii by capturing the dependencies between the output variables $(y_{i,m}^a, y_{i}^b, y_{i,m}^c)$.  Moreover, we cast each cQA subtask as a structured prediction problem in order to model the dependencies between output variables of the same type. Our intuition is that if two comments $c_m^i$ and $c_n^i$ in the same thread are similar, then they are likely to have the same labels for both subtask A and subtask C, i.e., $y_{i,m}^a \approx y_{i,n}^a$, and $y_{i,m}^c \approx y_{i,n}^c$. Similarly, if two pre-existing questions $q_i$ and $q_j$ are similar, they are also likely to have the same labels, i.e., $y_i^b \approx y_j^b$. 

\begin{figure*}[t!]
\centering
\scalebox{0.80}{
\begin{subfigure}{.45\textwidth}
  \centering
  \hspace*{-4mm}\includegraphics[width=1\linewidth]{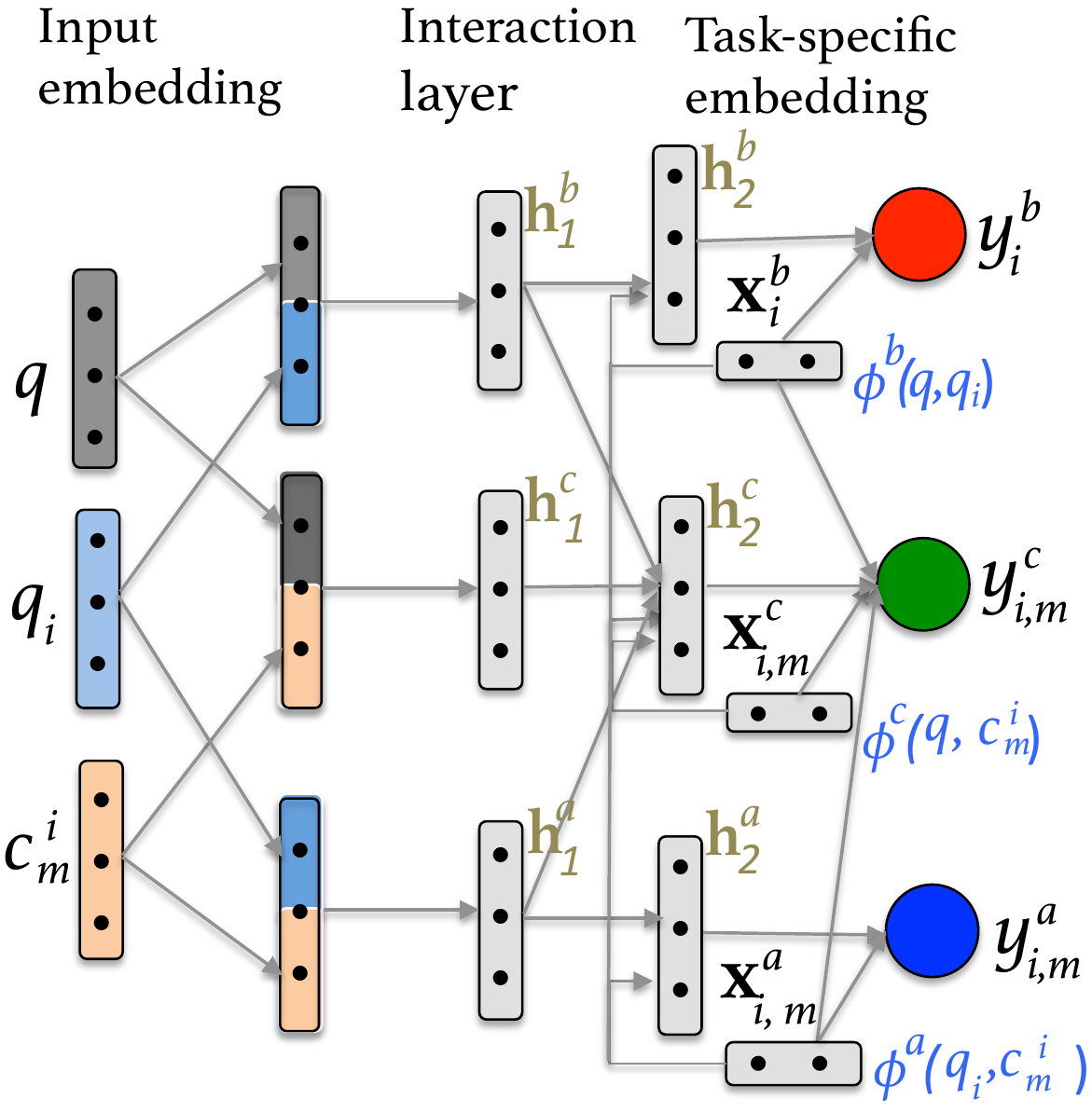}
  \caption{{\normalsize Our feed-forward neural networks}}
  \label{fig:dnn}
\end{subfigure}
\hspace{2em}
\begin{subfigure}{.55\textwidth}
  \centering
  \includegraphics[height=3in,width=2.5in]{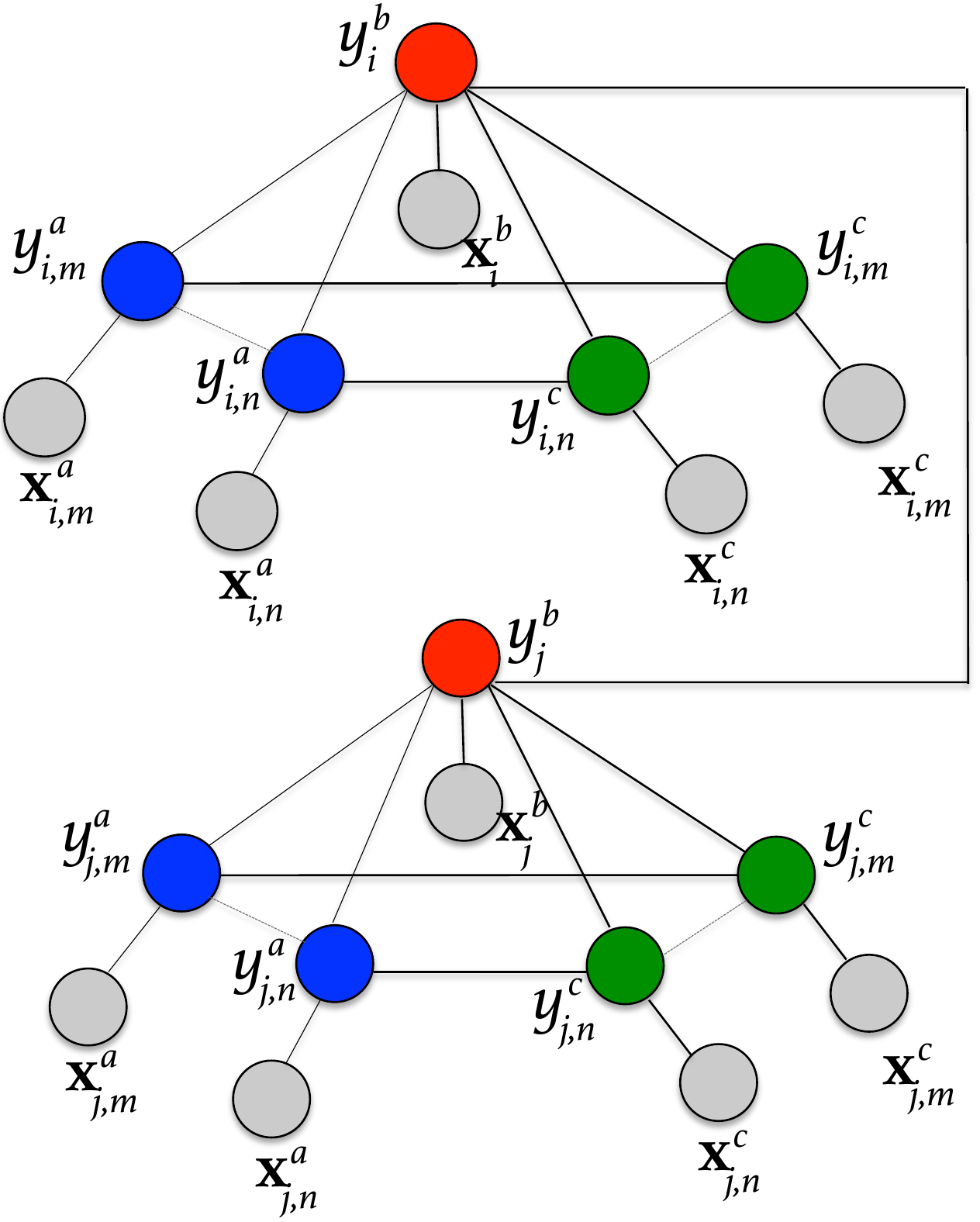}
  \caption{{\normalsize Our globally normalized joint conditional model}}
  \label{fig:crf}
\end{subfigure}}
\caption{Graphical representation of our cQA framework. On the left (a), we have three feed-forward neural networks to learn task-specific embeddings for the three cQA subtasks. On the right (b), a global conditional random field (CRF) models intra- and inter-subtask dependencies.}
\label{fig:models}
\vspace{-0.2em}
\end{figure*}

Our framework works in two steps. First, we use a DNN, specifically, a feed-forward NN, to learn task-specific embeddings for the three subtasks, i.e., output embeddings $\mathbf{x}_{i,m}^a$, $\mathbf{x}_{i}^b$ and $\mathbf{x}_{i,m}^c$ for subtasks A, B and C (Figure ~\ref{fig:dnn}). The DNN uses syntactic and semantic embeddings of the input elements, their interactions, and other similarity features between them and, as a by-product, learns the output embeddings for each subtask.

\noindent In the second step, a structured conditional model operates on subtask-specific embeddings from the DNNs and captures the dependencies between the subtasks, between existing questions, and between comments for an existing question (Figure~\ref{fig:crf}). Below, we describe the two steps in detail.

%% file: dnn-model.tex

Figure \ref{fig:dnn} depicts our complete neural framework for the three subtasks. The input is a tuple $(q, q_i, c_m^i)$ consisting of a new question $q$, a retrieved question $q_i$, and a comment $c_m^i$ from $q_i$'s answer thread. We first map the input elements to fixed-length vectors $(\mathbf{z}_{q}, \mathbf{z}_{q_i},\mathbf{z}_{c_m^i})$ using their syntactic and semantic \emph{embeddings}. Depending on the requirements of the subtasks, the network then models the interactions between the inputs  by passing their embeddings through non-linear hidden layers $\nu(\cdot)$. Additionally, the network also considers \emph{pairwise} similarity features $\phi(\cdot)$ between two input elements that go directly to the output layer, and also through the last hidden layer. The pairwise features together with the activations at the final hidden layer constitute the task-specific embeddings for each subtask $t$: $\mathbf{x}^t_{i}=[\nu^t(\cdot), \phi^t(\cdot)]$. 
The final layer defines a Bernoulli distribution for each subtask $t$ $\in$ $\{ a, b, c\}$:

\vspace{-1em}
\begin{equation}
p(y_{i}^t|q, q_i, c_m^i, \theta)= \Ber(y_{i}^t| \sig(\mathbf{w}_t^T \mathbf{x}^t_{i})) \label{loss}
\end{equation}

\noindent  where $\mathbf{x}^t_{i}$, $\mathbf{w}_t$, and $y_{i}^t$ are the task-specific embedding, the output layer weights, and the prediction variable for subtask $t$, respectively, and $\sig (\cdot)$ refers to the sigmoid function.

\noindent We train the models by minimizing the cross-entropy between the predicted distribution and the gold labels.
The main difference between the models is how they compute the task-specific embeddings $\mathbf{x}^t_{i}$ for subtask $t$.     

\paragraph{Neural Model for Subtask A.} The feed-forward network for subtask A is shown in the lower part of Figure \ref{fig:dnn}. To determine whether a comment $c_m^i$ is \emph{good} with respect to the thread question $q_i$, we model the interactions between $c_m^i$ and $q_i$ by merging their embeddings $\mathbf{z}_{c_m^i}$ and $\mathbf{z}_{q_i}$, and passing them through a hidden layer:  

\vspace{-0.5em}
\begin{equation}
\mathbf{h}_1^a =  f(U^a [\mathbf{z}_{q_i},\mathbf{z}_{c_m^i}]) 
\end{equation}

\noindent where $U^a$ is the weight matrix from the inputs to the first hidden units, $f$ is a non-linear activation function. The activations are then fed to a final subtask-specific hidden layer, which combines these signals with the pairwise similarity features $\phi^a(q_i, c_m^i)$. Formally, 

\vspace{-0.5em}
\begin{equation}
\mathbf{h}_2^a =  f(V^a [\mathbf{h}_{1}^a,\phi^a(q_i, c_m^i)]) 
\end{equation}

\noindent where $V^a$ is the weight matrix. The task-specific output embedding is formed by merging $\mathbf{h}_2^a$ and $\phi^a(q_i, c_m^i)$; $\mathbf{x}^a_{i,m}=[\mathbf{h}_2^a,\phi^a(q_i, c_m^i)]$.

\paragraph{Neural Model for Subtask B.} 
To determine whether an existing question $q_i$ is \emph{related} to the new question $q$, we model the interactions between $q$ and $q_i$ using their embeddings and pairwise similarity features similarly to subtask A. 

\noindent The upper part of Figure \ref{fig:dnn} shows the network. The transformation is defined as follows:

\vspace{-1em}
\begin{equation}
\mathbf{h}_1^b =  f(U^b [\mathbf{z}_{q}, \mathbf{z}_{q_i}]); \nonumber
\mathbf{h}_2^b =  f(V^b [\mathbf{h}_{1}^b,\phi^b(q, q_i)]) 
\end{equation}

\noindent where $U^b$ and $V^b$ are the weight matrices in the first and second hidden layer. The task-specific embedding is formed by $\mathbf{x}^b_{i}=[\mathbf{h}_2^b,\phi^b(q, q_i)]$.     

\paragraph{Neural Model for Subtask C.} The network for subtask C is shown in the middle of Figure \ref{fig:dnn}. To decide if a comment $c_m^i$ in the  thread of $q_i$ is \emph{relevant} to $q$, we consider how related $q_i$ is to $q$, and how useful $c_m^i$ is to answer $q_i$.  Again, we model the direct interactions between $q$ and $c_m^i$ using pairwise features $\phi^c(q, c_m^i)$ and a hidden layer transformation $\mathbf{h}_1^c =  f(U^c [\mathbf{z}_{q}, \mathbf{z}_{c^i_m}])$, where $U^c$ is a weight matrix. We then include a second hidden layer to combine the activations from different inputs and pairwise similarity features. Formally,  

\small
\vspace{-1em}
\begin{equation}
\mathbf{h}_2^c =  f(V^c [\mathbf{h}_{1}^a, \mathbf{h}_{1}^b, \mathbf{h}_{1}^c, \phi^a(q_i, c_m^i), \phi^b(q, q_i), \phi^c(q, c_m^i)]) \nonumber
\end{equation}
\normalsize

The final task-specific embedding for subtask C is formed as \small $\mathbf{x}^c_{i}=[\mathbf{h}_2^c,\phi^a(q_i, c_m^i), \phi^b(q, q_i), \phi^c(q, c_m^i)]$.\normalsize

%% file: crf-model.tex

One simple way to exploit the interdependencies between the subtask-specific embeddings $(\mathbf{x}_{i,m}^a$, $\mathbf{x}_{i}^b$, $\mathbf{x}_{i,m}^c)$ is to precompute the predictions for some subtasks (A and B), and then to use the predictions as features for the other subtask (C). However, as shown later in Section~\ref{sec:results}, such a \emph{pipeline} approach propagates errors from one subtask to the subsequent ones. A more robust way is to build a joint model for all subtasks. 

We could use the full DNN network in Figure~\ref{fig:dnn} to learn the classification functions for the three subtasks \emph{jointly} as follows:

\small
\vspace{-0.3cm}
\begin{equation}
p(y_{i,m}^a, y_i^b, y_{i,m}^c|\theta) = p(y_{i,m}^a|\theta_a) p(y_i^b|\theta_b) p(y_{i,m}^c|\theta_c) \label{eq:local_jnt}
\end{equation}
\normalsize

\noindent where $\theta = [\theta_a, \theta_b, \theta_c]$ are the model parameters.

However, this has two key limitations: \Ni it assumes conditional independence between the subtasks given the parameters;  \Nii the scores are normalized locally, which leads to the so-called \emph{label bias} problem~\cite{Lafferty01}, i.e., the features for one subtask would have no influence on the other subtasks.

Thus, we model the dependencies between the output variables by learning (globally normalized) node and edge factor functions that jointly optimize a global performance criterion. In particular, we represent the cQA setting as a large undirected graph $G$$=$$(V,E)$$=$$(V_a$$\cup$$V_b$$\cup$$V_c$, $E_{aa}$$\cup$$E_{bb}$$\cup$$E_{cc}$$\cup$$E_{ac}$$\cup$$E_{bc}$$\cup$$E_{ab})$. As shown in Figure~\ref{fig:crf}, the graph contains six subgraphs: $G_a$$=$$(V_a, E_{aa})$, $G_b$$=$$(V_b, E_{bb})$ and $G_c$$=$$(V_c, E_{cc})$ are associated with the three subtasks, while the bipartite subgraphs $G_{ac}$$=$$(V_a \cup V_c, E_{ac})$, $G_{bc}$$=$$(V_b \cup V_c, E_{bc})$ and $G_{ab}$$=$$(V_a \cup V_b, E_{ab})$ connect nodes across tasks.

We associate each node $u$ $\in$ $V_t$ with an input vector $\mathbf{x}_u$, representing the embedding for subtask $t$, and an output variable $y_u$, representing the class label for subtask $t$. Similarly, each edge $(u,v)$ $\in$ $E_{st}$ is associated with an input feature vector $\mu(\mathbf{x}_u,\mathbf{x}_v)$, derived  from the node-level features, and an output variable $y_{uv}$ $\in$ $\{1, 2, \cdots, L\}$, representing the state transitions for the pair of nodes.\footnote{To avoid visual clutter, the input features and the output variables for the edges are not shown in Figure~\ref{fig:crf}.} For notational simplicity, here we do not distinguish between comment and question nodes, rather we use $u$ and $v$ as general indices. We define the following joint conditional distribution:

\small
\vspace{-1em}
\begin{eqnarray}
p(\mathbf{y}|\theta, \mathbf{x}) = \frac{1}{Z(\theta, \mathbf{x})} \prod_{t \in \tau} \Big[ \prod_{u \in V_t} \psi_n(y_u|\mathbf{x}, \mathbf{w}_n^t) \Big] \nonumber \\ 
\hspace{0.5cm} \prod_{(s,t) \in \tau \times \tau} \hspace{0.2cm} \Big[ \hspace{-0.3cm} \prod_{(u,v) \in E_{st}}  \psi_e(y_{uv}|\mathbf{x}, \mathbf{w}_e^{st}) \Big] \label{eq:obj}
\end{eqnarray}
\normalsize

\noindent where $\tau = \{a,b,c\}$, $\psi_n (\cdot)$ and $\psi_e (\cdot)$ are node and edge \emph{factors}, respectively, and $Z(\cdot)$ is a global normalization constant. We use log-linear factors:

\vspace{-1em}
\small
\begin{eqnarray}
\psi_n(y_u|\mathbf{x}, \mathbf{w}_n^t) &=& \exp(\sigma(y_u, \mathbf{x})^T \mathbf{w}_n^{t}) \\
\psi_e(y_{uv}|\mathbf{x}, \mathbf{w}_e^{st}) &=& \exp(\sigma(y_{uv},\mathbf{x})^T \mathbf{w}_e^{st}) \label{eq:edgePot}
\end{eqnarray}
\normalsize

\noindent where $\sigma(\cdot)$ is a feature vector derived from the inputs and the labels.

This model is essentially a pairwise conditional random field \cite{Kevin12}. The global normalization allows CRFs to surmount the \emph{label bias} problem, allowing them to take long-range interactions into account. The objective in Equation \ref{eq:obj} is a convex function, and thus we can use gradient-based methods to find the global optimum. The gradients have the following form:

\small
\begin{eqnarray}
 f'(\mathbf{w}_n^t) &=& \sum_{u \in V_t} \sigma(y_u, \mathbf{x}) - \expect[\sigma(y_u, \mathbf{x})]  \\
 f'(\mathbf{w}_e^{st}) &=& \hspace{-0.5cm} \sum_{(u,v) \in E_{st}} \hspace{-0.2cm} \sigma(y_{uv}, \mathbf{x}) - \expect[\sigma(y_{uv},\mathbf{x})]
\end{eqnarray}
\normalsize

\noindent where $\expect[\phi(\cdot)]$ is the expected feature vector. 

\paragraph{Training and Inference.}

Traditionally, CRFs have been trained using offline methods like LBFGS \cite{Kevin12}. Online training using first-order methods such as stochastic gradient descent was proposed by \newcite{Vishwanathan:2006}. Since our DNNs are trained with the RMSprop online adaptive algorithm \cite{Tieleman12}, in order to compare our two models, we use RMSprop to train our CRFs as well.

For our CRF models, we use Belief Propagation, or BP, \cite{Pearl:1988} for inference. BP converges to an exact solution for trees. However, exact inference is intractable for graphs with loops. Despite this, \newcite{Pearl:1988} advocated for the use of BP in loopy graphs as an approximation. Even though BP only gives approximate solutions, it often works well in practice for loopy graphs \cite{Murphy99:LBP}, outperforming other methods such as mean field \cite{Weiss:01}.

\paragraph{Variations of Graph Structures.} 

A crucial advantage of our CRFs is that we can use arbitrary graph structures, which allows us to capture dependencies between different types of variables: \Ni \emph{intra-subtask}, for variables of the same subtask, e.g.,~$y_{i}^b$ and $y_{j}^b$ in Figure~\ref{fig:crf}, and \Nii \emph{across-subtask}, for variables of different subtasks.

For intra-subtask, we explore \emph{null} (i.e., no connection between nodes) and \emph{fully-connected} relations. For subtasks A and C, the intra-subtask connections are restricted to the nodes inside a thread, e.g., we do not connect $y_{i,m}^c$ and $y_{j,m}^c$ in Figure~\ref{fig:crf}.

For across-subtask, we explored three types of connections depending on the subtasks involved: \Ni \emph{null} or no connection between subtasks, \Nii \emph{1:1} connection for A-C, where the corresponding nodes of the two subtasks in a thread are connected, e.g., $y_{i,m}^a$ and $y_{i,m}^c$ in Figure~\ref{fig:crf}, and \Niii \emph{M:1} connection to B, where we connect all the nodes of C or A to the thread-level B node. Each configuration of intra- and across-connections yields a different CRF model. Figure~\ref{fig:crf} shows one such model for two threads each containing two comments, where all subtasks have fully-connected intra-subtask links, 1:1 connection for A-C, and M:1 for C-B and A-B.

%% file: features.tex

We have two types of features: 
\Ni \emph{input embeddings}, for $q$, $q_i$ and $c_m^i$, and
\Nii \emph{pairwise features}, for $(q,q_i)$, $(q,c^i_m)$, and $(q_i,c^i_m)$ --- see Figure~\ref{fig:dnn}.

\subsection{Input Embeddings}

We use three types of pre-trained vectors to represent a question ($q$ or $q_i$) or a comment ($c_m^i$): 

{\bf \textsc{Google Vectors.}}
300-dimensional embedding vectors,
trained on 100 billion words from Google News \cite{mikolov-yih-zweig:2013:NAACL-HLT}.
The embedding for a question (or comment) is the average of the word embeddings it is composed of.

{\bf \textsc{Syntax.}} We parse the question (or comment) using the Stanford neural parser \cite{socher-EtAl:2013:ACL2013}, and we use the final 25-dimensional vector produced internally as a by-product of parsing.

{\bf \textsc{QL Vectors.}} We use fine-tuned word embeddings pretrained on all the available in-domain Qatar Living data \cite{SemEval2016:task3:SemanticZ}.

\subsection{Pairwise Features}

We extract pairwise features for each of $(q,q_i)$, $(q,c^i_m)$, and $(q_i,c^i_m)$ pairs. These include:

{\bf \textsc{Cosines.}}
We compute cosines using the above vectors:
$\cos(q,q_i)$, $\cos(q,c^i_m)$ and $\cos(q_i,c^i_m)$.

{\bf \textsc{MT Features.}}
We use the following machine translation evaluation metrics:
(1)~\textsc{Bleu} \cite{Papineni:Roukos:Ward:Zhu:2002};
(2)~\textsc{NIST} \cite{Doddington:2002:AEM};
(3)~\textsc{TER} v0.7.25 \cite{Snover06astudy};
(4)~\textsc{Meteor} v1.4 \cite{Lavie:2009:MMA};
(5)~Unigram ~\textsc{Precision}; 
(6)~Unigram ~\textsc{Recall}.

{\bf \textsc{BLEU Components.}} We further use various components 
involved in the computation of \textsc{Bleu}:\footnote{\textsc{BLEU Features} and \textsc{BLEU Components} \cite{guzman-marquez-nakov:2016:P16-2,guzman-nakov-marquez:2016:SemEval} are ported from an MT evaluation framework \cite{guzman-EtAl:2015:ACL-IJCNLP,GuzmanJMN17} to cQA.}
$n$-gram precisions,
$n$-gram matches,
total number of $n$-grams ($n$=1,2,3,4),
lengths of the hypotheses and of the reference, 
length ratio between them,
and \textsc{Bleu}'s brevity penalty.

{\bf \textsc{Question-Comment Ratio.}}
(1)~question-to-comment count ratio in terms of senten\-ces/tokens/nouns/verbs/adjectives/adverbs/pronouns;
(2)~question-to-comment count ratio of words that are not in \textsc{word2vec}'s Google News vocabulary.

\subsection{Node Features}

{\bf \textsc{Comment Features.}} These include number of 
(1)~nouns/verbs/adjectives/adverbs/pronouns,
(2)~URLs/images/emails/phone numbers,
(3)~tokens/sentences,
(4)~positive/negative smileys,
(5)~single/double/triple exclamation/interrogation symbols,
(6)~interrogative sentences,
(7)~`thank' mentions,
(8)~words that are not in \textsc{word2vec}'s Google News vocabulary.
Also, (9)~average number of tokens, and
(10)~word type-to-token ratio.

{\bf \textsc{Meta Features.}}
(1)~is the person answering the question the one who asked it;
(2)~reciprocal rank of comment $c_m^i$ in the thread of $q_i$, i.e., $1/m$;
(3)~reciprocal rank of $c_m^i$ in the list of comments for $q$, i.e., $1$$/$$[m$$+$$10$$\times$$(i-1)]$;
and
(4)~reciprocal rank of question $q_i$ in the list for $q$, i.e., $1/i$.

%% file: data-setting.tex

We experiment with the data from SemEval-2016 Task 3 \cite{nakov-EtAl:2016:SemEval}. %
Consistently with our notation from Section~\ref{sec:approach}, it features three subtasks: subtask A (i.e.,~whether a comment $c_m^i$ is a good answer to the question $q_i$ in the thread), subtask B (i.e.,~whether the retrieved question $q_i$ is related to the new question $q$), and subtask C (i.e.,~whether the comment $c_m^i$ is a relevant answer for the new question $q$). 
Note that the two main subtasks we are interested in are B and C.

\vspace{-0.5em}
\paragraph{DNN Setting.}

We preprocess the data using min-max scaling. We use RMSprop\footnote{Other adaptive algorithms such as ADAM \cite{KingmaB14} or ADADELTA \cite{Zeiler12} were slightly worse.} 
for learning, with parameters set to the values suggested by \newcite{Tieleman12}. We use up to 100 epochs with patience of 25,
rectified linear units (ReLU) as activation functions, $l_2$ regularization on weights, and dropout \cite{Srivastava14a} of hidden units. See Table~\ref{best-dnn-setting} for more detail.

\begin{table}[tb!]
\scalebox{0.75}{\begin{tabular}{@{ }lccccc@{ }}
\toprule
& \bf batch & \bf dropout & \bf reg. str  & \bf inter. layer    & \bf task-spec. layer \\ 
\midrule
A & 16 & 0.3 & 0.001 & 10 & 125\\
B & 25 & 0.2 & 0.05 & 5  & 75 \\
C & 32 & 0.3 & 0.0001 & 15 & 50\\
\bottomrule
\end{tabular}}
\vspace{-0.6em}
\caption{Best setting for DNNs, as found on \textsc{dev}.}
\label{best-dnn-setting}
\vspace{-1em}
\end{table}

\vspace{-0.5em}
\paragraph{CRF Setting.}
For the CRF model, we initialize the node-level weights from the output layer weights of the DNNs, and we set the edge-level weights to 0.  
Then, we train using RMSprop with loopy BP.
We regularize the node parameters according to the best settings of the DNN: 0.001, 0.05, and 0.0001 for A, B, and C, respectively.

%% file: dnn-eval.tex

Table~\ref{tab:mainresults} shows the results for our individual DNN models (rows in boldface) for subtasks A, B and C on the \textsc{test} set. 

\noindent We report three ranking-based measures that are commonly accepted in the IR community: mean average precision (MAP), which was the official evaluation measure of SemEval-2016, average recall (AvgRec), and mean reciprocal rank (MRR).

For each subtask, we show two baselines and the results of the top-2 systems at SemEval.
The first baseline is a random ordering of the questions/comments, assuming no knowledge about the subtask. The second baseline keeps the chronological order of the comments for subtask A, of the question ranking from the IR engine for subtask~B, and both for subtask C.

\begin{table}[t!]
\centering
\footnotesize
\scalebox{0.9}{\begin{tabular}{lccc}
\toprule
\multicolumn{4}{l}{\bf Subtask A}\\
\midrule
\emph{System} & MAP & AvgRec & MRR\\
\midrule
Random order        & 52.80 & 66.52 & 58.71 \\
Chronological order & 59.53 & 72.60 & 67.83 \\
\midrule
ConvKN (second at SE-2016) & 77.66 & 88.05 & 84.93 \\
Kelp (best at SE-2016) & 79.19 & 88.82 & 86.42 \\
\bf DNN$_\mathbf{A}$ (subtask A network) & \bf 76.20 & \bf 86.52 & \bf 84.95\\
\toprule
\multicolumn{4}{l}{\vspace*{-2mm}}\\
\multicolumn{4}{l}{\bf Subtask B}\\
\midrule
\emph{System} & MAP & AvgRec & MRR\\
\midrule
Random order & 46.98 & 67.92 & 50.96 \\
IR order     & 74.75 & 88.30 & 83.79 \\
\midrule
ConvKN (second at SE-2016) & 76.02 & 90.70 & 84.64 \\
UH-PRHLT (best at SE-2016) & 76.70 & 90.31 & 83.02 \\
\bf DNN$_\mathbf{B}$ (subtask B network) & \bf 76.27 & \bf 90.27 & \bf 83.57 \\
\hline
DNN$_B$ + A gold labels & 76.10 & 89.96 & 83.62 \\
DNN$_B$ + C gold labels & 77.19 & 90.78 & 83.73 \\
DNN$_B$ + A and C gold labels & 77.12 & 90.71 & 83.73 \\
\toprule
\multicolumn{4}{l}{\vspace*{-2mm}}\\
\multicolumn{4}{l}{\bf Subtask C}\\
\midrule
\emph{System} & MAP & AvgRec & MRR\\\hline
Random order           & 15.01 &  11.44 &  15.19 \\
IR+Chron. order & 40.36 &  45.97 &  45.83 \\
\midrule
Kelp (second at SE-2016) & 52.95 & 59.27 & 59.23 \\
SUper team (best at SE-2016) & 55.41 & 60.66 & 61.48 \\
\bf DNN$_\mathbf{C}$ (subtask C network) & \bf 54.24 & \bf 58.30 & \bf 61.47 \\
\hline
DNN$_C$ + A gold labels & 61.14 & 66.67 & 66.86 \\
DNN$_C$ + B gold labels & 56.29 & 61.11 & 62.67 \\
DNN$_C$ + A and B gold labels & 63.49 & 71.16 & 68.19	\\
\bottomrule
\end{tabular}
}
\vspace{-0.5em}
\caption{Results for our DNN models on all cQA subtasks, compared to the top-2 systems from SemEval-2016 Task 3.
Inter-subtask dependencies are explored using gold output labels.}
\label{tab:mainresults}
\vspace{-1em}
\end{table}

We can see that the individual DNN models for subtasks B and C are very competitive, falling between the first and the second best at SemEval-2016.
For subtask A, our model is weaker, but, as we will see below, it can help improve the results for subtasks B and C, which are our focus here.

Looking at the results for subtask C, we can see that sizeable gains are possible when using gold labels for subtasks A and B as features to DNN$_C$, e.g.,~adding gold~A labels yields +6.90 MAP points.

\noindent Similarly, using gold labels for subtask B adds +2.05 MAP points absolute. 
Moreover, the gain is cumulative: using the two gold labels together yields +9.25 MAP points. The same behavior is observed for the other evaluation measures. 
Of course, as we use gold labels, this is an upper bound on performance, but it justifies our efforts towards a joint multitask learning model.

%% file: joint-eval.tex

\begin{table*}[t]
\centering
\small
\scalebox{0.9}{\begin{tabular}{clllll}
\toprule
\emph{\#} & \emph{System} & \emph{Comments} & \emph{MAP} ($\Delta$) & \emph{AvgRec} ($\Delta$) & \emph{MRR} ($\Delta$)\\
\midrule
1 & DNN$_C$ & Subtask C network & 54.24 & 58.30 & 61.47 \\
\midrule
2 & DNN$_{C+PA}$    & DNN$_C$ with A predicted labels & 55.21 \footnotesize{(+0.97)}& 58.36 \footnotesize{(+0.06)}& 62.69 \footnotesize{(+1.22)}\\
3 & DNN$_{C+PB}$    & DNN$_C$ with B predicted labels & 54.17 \footnotesize{(-0.04)}& 58.17 \footnotesize{(-0.13)}& 62.55 \footnotesize{(+1.08)}\\
4 & DNN$_{C+PA+PB}$ & DNN$_C$ with A and B predicted labels & 55.11 \footnotesize{(+0.90)}& 58.69 \footnotesize{(+0.39)}& 60.10 \footnotesize{(-1.37)}\\
\midrule
5 & CRF$_{AC}$ & CRF with A-C connections  & 55.42 \footnotesize{(+1.18)} & 58.69 \footnotesize{(+0.39)}& 63.25 \footnotesize{(+1.78)}\\
6 & CRF$_{BC}$ & CRF with B-C connections & 55.20 \footnotesize{(+0.96)} & 58.87 \footnotesize{(+0.57)}& 62.30 \footnotesize{(+0.83)}\\
7 & CRF$_{ACBC}$ & CRF with A-C and B-C connections & {\bf 56.00} \footnotesize{(+1.76)} & 60.20 \footnotesize{(+1.90)}& {\bf 63.25} \footnotesize{(+1.78)}\\
8 & CRF$_{all}$ & CRF with all pairwise connections & 
55.81 \footnotesize{(+1.57)} & 60.15 \footnotesize{(+1.85)} & 62.68 \footnotesize{(+1.21)}\\
\midrule
9 & CRF$_{ACBC,C^f}$ & CRF$_{ACBC}$ with fully connected C & 55.73 \footnotesize{(+1.49)} & 59.77 \footnotesize{(+1.47)} & 62.80 \footnotesize{(+1.33)}\\
10 & CRF$_{ACBC,A^fC^f}$ & CRF$_{ACBC}$ with fully connected A and C & 55.54 \footnotesize{(+1.30)}& 59.86 \footnotesize{(+1.56)}& 62.54 \footnotesize{(+1.07)}\\
11 & CRF$_{ACBC,B^fC^f}$ & CRF$_{ACBC}$ with fully connected B and C & 55.67 \footnotesize{(+1.43)} & {\bf 60.22} \footnotesize{(+1.92)} & 62.80 \footnotesize{(+1.33)}\\
12 & CRF$_{ACBC,f}$ &  CRF$_{ACBC}$ with all layers fully connected & 55.81 \footnotesize{(+1.57)} & 60.15 \footnotesize{(+1.85)} &  {\bf 63.25} \footnotesize{(+1.78)}\\
\bottomrule
\end{tabular}
}
\vspace{-0.6em}
\caption{Performance of the \emph{pipeline} and of the joint learning models on subtask C.
The best results for each measure are in bold, and the gains over the single neural network (DNN$_C$) are shown in parentheses.}
\label{tab:jointresultsC}
\vspace{-0em}
\end{table*}

Below we discuss the evaluation results for the joint model. We focus on subtasks B and C, which are the main target of our study.

\paragraph{Results for Subtask C.}

Table~\ref{tab:jointresultsC} compares several variants of the CRF model for joint learning, which we described in Section~\ref{subsec:crf-model} above.

Row 1 shows the results for our individual DNN$_C$ model.
The following rows 2--4 present a \emph{pipeline} approach, where we first predict labels
for subtasks A and B and then we add these predictions as features to DNN$_C$. This is prone to error propagation, and improvements are moderate and inconsistent across the evaluation measures.

The remaining rows correspond to variants of our CRF model with different graph structures.
Overall, the improvements over DNN$_C$ are more sizeable than for the pipeline approach (with one single exception out of 24 cases); they are also more consistent across the evaluation measures, and the improvements in MAP over the baseline range from +0.96 to +1.76 points absolute.
 
Rows 5--8 show the impact of adding connections to subtasks A and B when solving subtask C (see Figure~\ref{fig:crf}).
Interestingly, we observe the same pattern as with the gold labels: 
the A-C and B-C connections help individually and in combination, with A-C being more helpful.
Yet, further adding A-B does not improve the results (row~8). 

Note that the locally normalized joint model in Eq. \ref{eq:local_jnt} yields much lower results than the globally normalized CRF$_{all}$ (row~8): 54.32, 59.87, and 61.76 in MAP, AvgRec and MRR (figures not included in the table for brevity). This evinces the problems with the conditional independence assumption and the local normalization in the model.

Finally, rows 9--12 explore variants of the best system from the previous set (row 7), which has connections between subtasks only.
Rows 9--12 show the results when using subgraphs for A, B and C that are fully connected (i.e.,~for all pairs).

\noindent We can see that none of these variants yields improvements over the model from row 7,
i.e.,~the fine-grained relations between comments in the threads and between the different related questions do not seem to help solve subtask C in the joint model. 
Note that our scores from row 7 are better than the best results achieved by a system at SemEval-2016 Task 3 subtask C: 56.00 vs. 55.41 on MAP, and 63.25 vs. 61.48 on MRR.

\paragraph{Results for Subtask B.}

\begin{table*}[t]
\centering
\hspace*{-3mm}
\footnotesize
\scalebox{0.9}{\begin{tabular}{clllllllll}
\toprule
\emph{\#} & \emph{System} & \emph{Comments} & \emph{MAP} & \emph{AvgRec} & \emph{MRR} & \emph{Acc} & \emph{P} & \emph{R} & \emph{F$_1$}\\
\midrule
1 & DNN$_B$ & Subtask B network & 76.27 & 90.27 & 83.57 & 76.39 & \bf 89.53 & 33.05 & 48.28 \\\hline
2 & DNN$_{B+PA}$    & DNN$_B$ with A predicted labels & 76.08 & 89.99 & 83.38 & 77.40 & 86.41 & 38.20 & 52.98 \\
3 & DNN$_{B+PC}$    & DNN$_B$ with C predicted labels & 76.33 & 90.38 & 83.62 & 77.40 & 83.19 & 40.34 & 54.34 \\
4 & DNN$_{B+PA+PC}$ & DNN$_B$ with A and C predicted labels & 76.43 & 90.34 & 83.62 & 77.11 & 78.74 & 42.92 & 55.56 \\
\midrule
5 & CRF$_{B^f}$ & CRF with fully connected B & 76.41 & 90.34 & 83.81 & 77.00 & 84.62 & 37.76 & 52.23\\\hline 
6 & CRF$_{ACBC,B^f}$ & CRF$_{ACBC}$ with fully connected B & \bf 76.89 & 90.87 & 84.19 & 77.86 & 76.00 & \bf 48.93 & \bf 59.53 \\
7 & CRF$_{ACBC,A^fB^f}$ & CRF$_{ACBC}$ with fully connected A and B & 76.51 & 90.64 & 84.19 & 78.29 & 83.47 & 43.35 & 57.06 \\
8 & CRF$_{ACBC,B^fC^f}$ & CRF$_{ACBC}$ with fully connected B and C & 76.87 & \bf 90.96 & 84.44 & 77.86 & 78.68 & 45.92 & 58.00 \\
9 & CRF$_{ACBC,f}$ &  CRF$_{ACBC}$ with all layers fully connected & 76.25 & 90.38 & \bf 84.62 & \bf 78.57 & 81.20 & 46.35 & 59.02 \\
\bottomrule
\end{tabular}
}
\caption{Performance of the \emph{pipeline} and of the joint models on subtask B (best results in boldface).} 
\label{tab:jointresultsB}
\end{table*}

Next, we present in Table~\ref{tab:jointresultsB} similar experiments, but this time with subtask B as the target, and we show some more measures (accuracy, precision, recall, and F$_1$). 

Given the insights from Table~\ref{tab:mainresults} (where we used gold labels), we did not expect to see much improvements for subtask B.
Indeed, as rows \hbox{2--4} show, using the pipeline approach, the IR measures are basically unaltered.
However, classification accuracy improves by almost one point absolute, recall is also higher (trading for lower precision), and F$_1$ is better by a sizeable margin.

Coming to the joint models (rows 6--9), we can see that the IR measures improve consistently over the pipeline approach, even though not by much. The effect on accuracy-P-R-F$_1$ is the same as observed with the pipeline approach but with larger differences.\footnote{Note that we have a classification approach, which favors accuracy-P-R-F$_1$; if we want to improve the ranking measures, we should optimize for them directly.} 
In particular, accuracy improves by more than two points absolute, and recall increases, which boosts F$_1$ to almost 60.

Row 5 is a special case where we only consider subtask B, but we do the learning and the inference over the set of ten related questions, exploiting their relations.
This yields a slight increase in all measures; more importantly, it is crucial for obtaining better results with the joint models.

Rows 6--9 show results for various variants of the A-C and B-C architecture with fully connected B nodes, playing with the fine-grained connection of the A and C nodes.
The best results are in this block, with increases over DNN$_B$ in MAP (+0.61), AvgRec (+0.69) and MRR (+1.05),
and especially in accuracy (+2.18) and F$_1$ (+11.25 points).
This is remarkable given the low expectation we had about improving subtask B. 

Note that the best architecture for subtask C from Table~\ref{tab:jointresultsC} (A-C and B-C with no fully connected B layer) does not yield good results for subtask B.

\noindent We speculate that subtask B is overlooked by the architecture, which has many more connections and parameters on the nodes for subtasks A and C (ten comments are to be classified for both subtask A and C, while only one decision is to be made for the related question B).

Finally, note that our best results for subtask B are also slightly better than those for the best system at SemEval-2016 Task 3, especially on MRR.